\documentclass[acmtog]{acmart}
\usepackage{booktabs} % For formal tables

\usepackage[ruled]{algorithm2e} % For algorithms

\usepackage{xcolor}

\SetAlFnt{\small}
\SetAlCapFnt{\small}
\SetAlCapNameFnt{\small}
\SetAlCapHSkip{0pt}

\acmJournal{TOG}
\usepackage{enumitem}
\usepackage{multirow}
\begin{document}

\setcopyright{cc}
\setcctype{by}
\acmJournal{TOG}
\acmYear{2026}
\acmVolume{45}
\acmNumber{4}
\acmArticle{68}
\acmMonth{7}
\acmDOI{10.1145/3811399}

% Title portion
\title{SegviGen: Repurposing 3D Generative Model for Part Segmentation}

% DO NOT ENTER AUTHOR INFORMATION FOR ANONYMOUS TECHNICAL PAPER SUBMISSIONS TO SIGGRAPH 2019!
\author{Lin Li}
\authornote{Equal contribution}
\orcid{0009-0002-3113-416X}
\affiliation{%
  \institution{Renmin University of China}
  \city{Beijing}
  \country{China}}
\email{ruclilin@163.com}

\author{Haoran Feng}
\authornotemark[1]
\orcid{0009-0001-0770-361X}
\affiliation{%
  \institution{Tsinghua University}
  \city{Shenzhen}
  \country{China}}

\author{Zehuan Huang}
\authornote{Project lead}
\orcid{0009-0002-1883-0777}
\affiliation{%
  \institution{Beihang University}
  \city{Beijing}
  \country{China}}

\author{Haohua Chen}
\orcid{0009-0005-4631-0172}
\affiliation{%
  \institution{Beihang University}
  \city{Beijing}
  \country{China}}

\author{Wenbo Nie}
\orcid{0009-0008-7268-994X}
\affiliation{%
  \institution{Beijing Jiaotong University}
  \city{Beijing}
  \country{China}}

\author{Shaohua Hou}
\orcid{0009-0008-1172-940X}
\affiliation{%
  \institution{Beihang University}
  \city{Beijing}
  \country{China}}

\author{Keqing Fan}
\orcid{0009-0009-1032-2971}
\affiliation{%
  \institution{Beihang University}
  \city{Beijing}
  \country{China}}

\author{Pan Hu}
\orcid{0009-0006-8151-2279}
\affiliation{%
  \institution{Beihang University}
  \city{Beijing}
  \country{China}}

\author{Sheng Wang}
\orcid{0009-0004-1182-2453}
\affiliation{%
  \institution{Bambu Lab}
  \city{Shenzhen}
  \country{China}}

\author{Buyu Li}
\orcid{0000-0001-5442-4667}
\affiliation{%
  \institution{Bambu Lab}
  \city{Shenzhen}
  \country{China}}

\author{Lu Sheng}
\authornote{Corresponding author}
\orcid{0000-0002-8525-9163}
\affiliation{%
  \institution{Beihang University}
  \city{Beijing}
  \country{China}}
\email{lsheng@buaa.edu.cn}

\renewcommand{\shortauthors}{Li et al.}

\begin{abstract}
We introduce \textit{SegviGen}, a framework that repurposes native 3D generative models for 3D part segmentation. 
Existing pipelines either lift strong 2D priors into 3D via distillation or multi-view mask aggregation, 
often suffering from cross-view inconsistency and blurred boundaries, 
or explore native 3D discriminative segmentation, 
which typically requires large-scale annotated 3D data and substantial training resources.
In contrast, \textit{SegviGen} leverages the structured priors encoded in pretrained 3D generative model to induce segmentation through distinctive part colorization, establishing a novel and efficient framework for part segmentation.
Specifically, \textit{SegviGen} encodes an input 3D asset and predicts part-indicative colors on active voxels of a geometry-aligned reconstruction. 
It supports interactive part segmentation, full segmentation, and full segmentation with 2D guidance in a unified framework.
%
% Extensive experiments show that \textit{SegviGen} outperforms prior state of the art by approximately 40\% on interactive part segmentation and 15\% on full segmentation, while using only 0.32\% of the training data.
Extensive experiments show that \textbf{\textit{SegviGen} improves over the prior state of the art by 40\% on interactive part segmentation and by 15\% on full segmentation, while using only 0.32\% of the training data.}
This undoubtedly demonstrates that pretrained 3D generative priors transfer effectively to 3D part segmentation, enabling strong performance with limited supervision.
Code and pretrained weights are publicly available at https://github.com/Nelipot-Lee/SegviGen.
\end{abstract}
% \ccsdesc[300]{Computing methodologies~Artificial intelligence}

%
% End generated code
%

% \keywords{3D Segmentation, 3D Generative Model}

\begin{teaserfigure}
  \includegraphics[width=\textwidth]{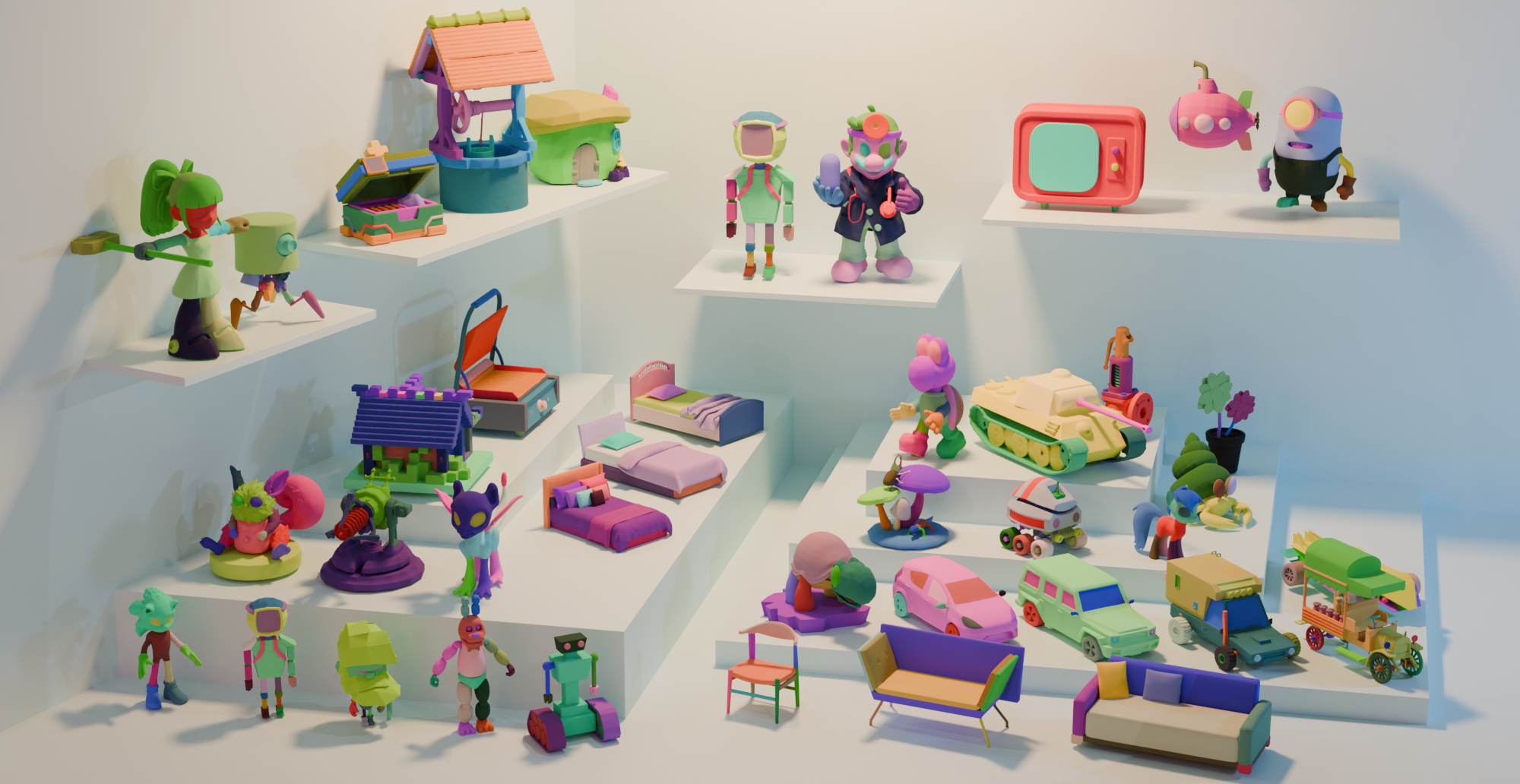}
  \caption{\textit{SegviGen} enables diverse and accurate 3D part segmentation by leveraging priors from large-scale 3D generative models. With substantially less training data, it produces high-fidelity segmentation results with sharp part boundaries and strong generalization across object categories.}
  \label{fig:teaser}
\end{teaserfigure}

\maketitle

% teaser % lin
% 3D 打印视频
% 渲染视频
\section{INTRODUCTION}

Part segmentation provides explicit part-level structures of 3D assets, serving as a core primitive for 3D content creation pipelines and offering fundamental 3D perception capabilities for spatial intelligence.
It enables a wide range of downstream applications, including part-level editing, animation rigging, and industrial uses such as 3D printing.
However, existing methods often fall short in segmentation quality, producing erroneous regions and imprecise boundaries that limit their practical usability.

To this end, one line of work attempts to transfer the comprehensive 2D segmentation priors to 3D via 2D-to-3D lifting.
Methods such as SAMPart3D~\cite{sampart3d} optimize 3D segmentation via 2D-to-3D distillation, 
but incur substantial computational and time overhead, and often yield blurry boundaries.
In parallel, another set of methods~\cite{segment3d,pointsam,geosam2} applies SAM~\cite{sam,sam2,sam3} to obtain 2D masks of multi-view projected images, which are then back-projected and fused into 3D masks.
However, these multi-view pipelines incur substantial runtime overhead, are sensitive to view coverage, and the back-projection and fusion step often introduces cross-view inconsistencies and imprecise boundaries.

% su du kuai
Recently, another line of work~\cite{p3sam,partsam} moves toward native 3D part segmentation so as to remedy the inherent shortcomings of the aforementioned methods that leverage 2D segmentation priors.
%
% These methods predict segmentation parts directly in the native 3D space, explicitly enforce semantic and structural consistency, and are more efficient at inference.
%
However, it is a typical requirement to collect large-scale training datasets with curated 3D part annotations, 
where fine-grained annotations are costly and inconsistent across sources in granularity, hierarchy, and boundary definitions.

%
% 总结，2D的问题，mismatch，3D的问题，然后本质promising是纯三维，41行，我们发现能够物体结构，外观这里，分割，prior
% Overall, existing approaches face a fundamental trade-off between annotation and training cost and the structural artifacts caused by lifting 2D cues into 3D.
% %
% We attribute these artifacts to the inherent mismatch between 2D priors and native 3D structure,
% %
% motivating us to instead exploit a native 3D structural prior in place of lifted 2D cues.
% In summary, the first line of methods suffers from a mismatch between 2D priors and 3D structure, while the second relies on costly training from scratch.
%
Therefore, a more promising approach is to leverage a prior model that encodes both 3D structure and texture to perform segmentation.
In particular, 3D generative models trained on large-scale unannotated 3D textured assets internalize rich part-level structure and texture patterns, providing a strong 3D prior over geometry and appearance. 
Such priors encourage part segmentation with sharper boundaries, while reducing reliance on dense part annotations and extensive task-specific training.
This motivates us to ask:
\textit{How can 3D generative priors be effectively transferred to part-level 3D segmentation to improve quality and data efficiency?}

Motivated by this perspective, we propose \textit{SegviGen}, a generative framework for 3D part segmentation that leverages the rich 3D structural and textural knowledge encoded in large-scale 3D generative models.
Specifically, we formulate part segmentation as a colorization task that leverages the full capacity of 3D generative models.
%
% \textit{SegviGen} encodes the input 3D asset into a latent representation and uses it, together with the task embedding and query points, to condition the denoising process.
%
The model is trained to predict part-indicative colors, along with reconstructing the underlying geometry.
This formulation naturally accommodates additional conditioning signals, enabling \textit{SegviGen} to flexibly support interactive part segmentation, full segmentation, and 2D segmentation map–guided full segmentation under a unified architecture. 
Notably, 2D segmentation map-guided full segmentation allows users to customize the decomposition through a 2D segmentation map.
%
% , providing a practical mechanism for customized 3D part parsing.
% 写清楚哪几个setting

Qualitative and quantitative results show that \textit{SegviGen} consistently surpasses the prior state-of-the-art, P3-SAM~\cite{p3sam}, while using only 0.32\% of the training data. % 写清楚啥方法。
On interactive part segmentation, it achieves the best performance across all metrics on PartObjaverse-Tiny~\cite{partobjaverse_tiny} and PartNeXT~\cite{partnext}, with a 40\% gain in IoU@1, an important metric that reflects the model’s single-click accuracy. 
% 体现这个指标的挑战性 
%
On full segmentation without guidance, \textit{SegviGen} outperforms the best baseline by 15\% in overall IoU, averaged across datasets.
Our main contributions are summarized as follows:
\begin{itemize}[leftmargin=*, topsep=2pt, itemsep=1pt, parsep=0pt, partopsep=0pt]
    \item We propose \textit{SegviGen}, a unified multi-task framework for 3D part segmentation that effectively exploits the structural and textural priors encoded in pretrained 3D generative models, enabling accurate and efficient segmentation.
    \item We reformulate 3D segmentation as part-wise colorization, where \textit{SegviGen} predicts the colors of actiave voxel as part labels in a single generative process.
    \item Extensive experiments show that \textit{SegviGen} outperforms the prior state of the art by 40\% on interactive part segmentation and 15\% on full segmentation, using only 0.32\% of the training data, highlighting the effectiveness of transferring 3D generative priors to part segmentation.
\end{itemize}

\section{Related Work}

\subsection{3D Part Segmentation}
% \paragraph{3D Part Segmentation}
% Traditional 3D part segmentation
Traditional 3D part segmentation is typically cast as supervised semantic labeling on points or faces, using fixed part taxonomies provided by curated 3D segmentation datasets~\cite{partnet,3dmeshsegmentation,scannet,pointnet}.
Concretely, these methods~\cite{pointnet,pointtransformerv2,pointtransformerv3,pointpromptraining,meshcnn,meshtransformer} typically combine a 3D feature encoder with a segmentation head to predict dataset-specific part IDs.
However, the closed-world nature of both the label space and the training data limits generalization, making it difficult to transfer to unseen object categories or arbitrary, non-canonical part decompositions.

To alleviate this generalization bottleneck, recent works exploit 2D foundation models as transferable priors~\cite{clip,glip,sam,sam2,dino,dinov2} for 3D part segmentation.
A common strategy adopts a render-and-lift pipeline: it segments multi-view renderings with promptable 2D models and then projects and fuses the masks back onto the 3D surface~\cite{segmentanymesh,sam3d,sampro3d,partslip++,zerops}.
%
% Despite being straightforward, this pipeline is often limited by incomplete view coverage and cross-view inconsistencies, which can lead to imprecise or blurred part boundaries after 2D-to-3D aggregation. 
%
Another line leverages distillation or feature projection to supervise 3D predictors with transferred 2D representations or pseudo-labels~\cite{partdistill,3dpartsegvia2dfea}. 
However, thest pipelines inherit the 2D--3D domain gap and multi-view alignment issues, and typically entails longer optimization and training cycles.

Recognizing the scalability and reliability issues of 2D-to-3D lifting, recent studies have shifted toward native feed-forward 3D segmentation that predicts masks directly on 3D representations at inference time.
Representative efforts for open-world part segmentation include training queryable 3D predictors with automatically curated supervision~\cite{find3d}, 
learning continuous part-aware 3D feature fields for direct decomposition~\cite{partfield}, 
and prompt-guided 3D mask prediction models~\cite{pointsam}, 
with more recent large-scale native 3D part segmentation models such as P3-SAM~\cite{p3sam} and PartSAM~\cite{partsam} further scaling training on millions of shape--part pairs. 
Despite encouraging progress, 
these native 3D approaches are fundamentally bottlenecked by the availability of large-scale, high-quality 3D part annotations, 
and the inconsistency of part taxonomies and granularity across datasets often introduces supervision mismatch, ultimately weakening cross-domain generalization.

\begin{figure*}
  \centering
  \includegraphics[width=\linewidth]{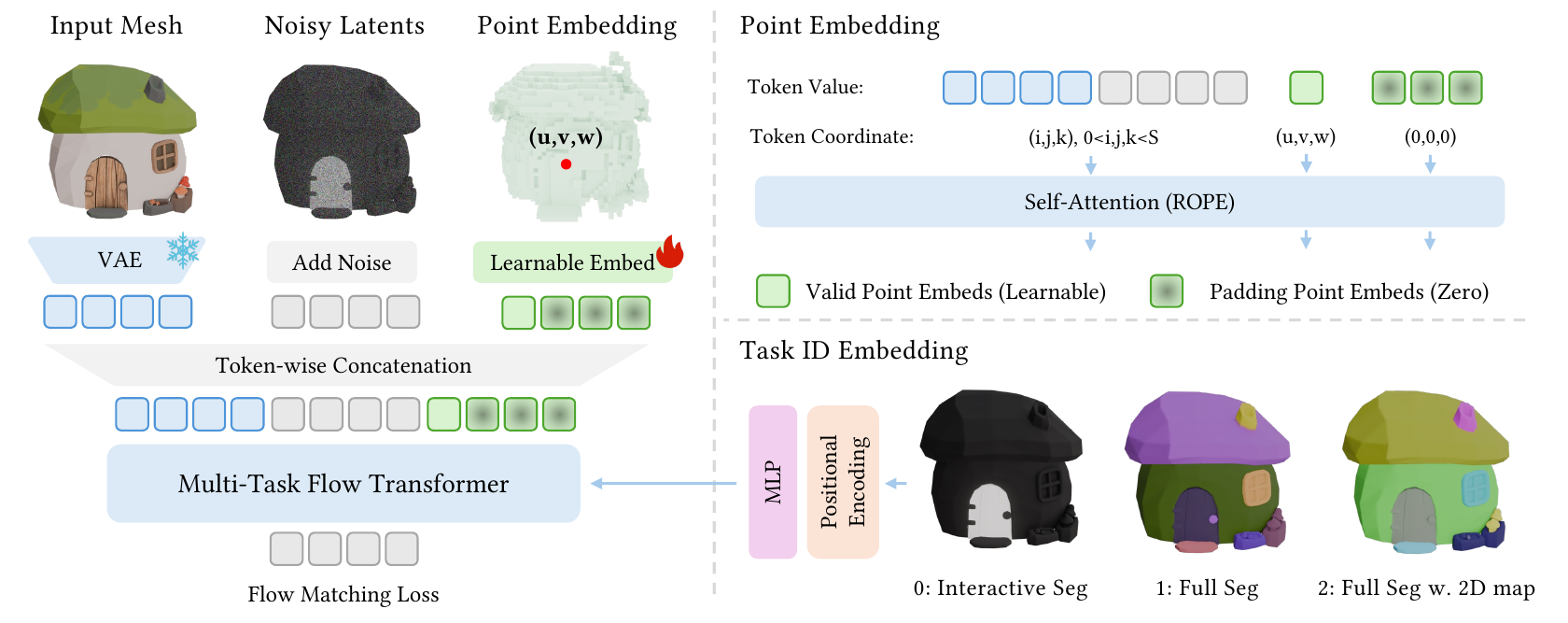}
  \caption{
  Pipeline of \textit{SegviGen}. 
  We reformulate 3D part segmentation as a conditional colorization task. 
  During training, given a 3D mesh and its part-color ground truth, we encode both with a pretrained 3D VAE, add noise to the ground-truth latent, and then concatenate the geometry latent, noisy color latent, and point-condition tokens to form the final latent input.
  Conditioned on the sampled timestep and a task embedding, the multi-task flow transformer predicts the noise residual for flow-matching training.
  }
  \label{fig:pipeline}
\end{figure*}

% \paragraph{3D Generative Model} 
\subsection{3D Generative Model}
The rapid progress of diffusion-based generative modeling~\cite{ho2020ddpm,song2020ddim}, together with the emergence of large-scale, high-quality 3D data collections~\cite{deitke2023objaverse,deitke2024objaversexl}, has catalyzed a wave of 3D generative methods~\cite{liu2024one2345,liu2023syncdreamer,long2024wonder3d,hong2023lrm,tang2025lgm,huang2024epidiff,zhang2024clay,wu2024unique3d,li2024craftsman,wen2024ouroboros3d,xu2024instantmesh,voleti2025sv3d,wang2024crm,liu2024one2345++,wu2024direct3d,zhao2024michelangelo,roessle2024l3dg,wu2024blockfusion,meng2024lt3sd,liu2024part123,dong2025tela,chen2024meshxlneuralcoordinatefield,chen2024meshanythingartistcreatedmeshgeneration,wang2024llamameshunifying3dmesh,hao2024meshtronhighfidelityartistlike3d,he2024neurallightrigunlockingaccurate,gao2025meshartgeneratingarticulatedmeshes,zhao2025deepmeshautoregressiveartistmeshcreation,wei2025octgptoctreebasedmultiscaleautoregressive,li2025step1x3dhighfidelitycontrollablegeneration,ye2025shapellmomninativemultimodalllm}.
A prevalent route builds 3D assets through a 2D-to-3D pipeline: models first synthesize multi-view imagery and subsequently reconstruct the underlying 3D geometry and appearance from these views
~\cite{liu2023syncdreamer,long2024wonder3d,tang2025lgm,wen2024ouroboros3d,xu2024instantmesh,wang2024crm,voleti2025sv3d,huang2024mvadapter,qu2025deocc1to33ddeocclusionsingle,huang2025stereogsmultiviewstereovision}, yet view-to-view discrepancies in the synthesized images can propagate and degrade the final 3D quality.

In contrast, a growing family of native 3D generative models learns directly in 3D latent spaces, typically pairing a variational autoencoder~\cite{kingma2013vae} with a diffusion transformer (DiT)~\cite{peebles2023dit} to perform denoising over compact latents
~\cite{zhang2024clay,li2024craftsman,wu2024direct3d,zhao2024michelangelo,li2025triposg,chen2025ultra3defficienthighfidelity3d,dong2025morecontextuallatents3d,zhao2025assemblerscalable3dassembly,tang2025efficientpartlevel3dobject,lin2025partcrafterstructured3dmesh,wu2025dipodualstateimagescontrolled,wu2025direct3ds2gigascale3dgeneration,li2025triposghighfidelity3dshape,trellis,li2025craftsman3dhighfidelitymeshgeneration,trellis2}.
By learning to generate in a compact yet expressive 3D latent space, these models encode rich structural and texture knowledge across large-scale 3D assets, providing a strong transferable prior for downstream 3D part segmentation.
In particular, TRELLIS2~\cite{trellis2} introduces a field-free structured latent via an omni-voxel sparse voxel representation (O-Voxel) that jointly models geometry and appearance, enabling efficient generation with sharp, high-frequency textures that better preserve fine-grained part boundaries for 3D segmentation.
\section{METHOLODOGY}

We propose \textit{SegviGen}, a unified multi-task framework for 3D part segmentation that supports three practical settings: interactive part-segmentation, full segmentation, and full segmentation with 2D guidance.
To leverage the prior knowledge encoded in a pretrained 3D generative model, we cast 3D segmentation as a colorization problem.
%
% Specifically, we encode the input 3D asset into a compact latent that conditions generation, and optionally augment it with user interactions or a 2D segmentation map.
%
Conditioned on these inputs, the model reconstructs the 3D asset while predicting colors for active voxels in the structured 3D representation, where each color corresponds to an individual part, yielding the final segmentation.
Below, we begin by describing the underlying 3D generative model (Sec.~\ref{sec:3d_generative_model}), followed by our task reformulation (Sec.~\ref{sec:task_define}), and then detail the overall pipeline (Sec.~\ref{sec:pipeline}).

\subsection{Preliminary: Structured-Latent 3D Generative Model}
\label{sec:3d_generative_model}

Recent work~\cite{trellis2} organizes each textured 3D asset into a sparse set of active voxels on a regular grid, where every active voxel stores geometry and texture features aligned in 3D.
%.
% This formulation leverages a flexible dual-grid construction to robustly handle arbitrary topology while encoding physically-based material attributes jointly with geometry for faithful appearance modeling.
%
Given the sparse omni-voxel representation, a Sparse Compression VAE (SC-VAE) maps each voxelized asset feature tensor $\mathbf{x}$ to a compact structured latent $\mathbf{z}_1=E_{\phi}(\mathbf{x})$ and reconstructs it via $\hat{\mathbf{x}}=D_{\theta}(\mathbf{z}_1)$, yielding an expressive yet highly compressed 3D latent space.
On top of these latents, a conditional flow-matching generator learns a time-dependent vector field $\mathbf{v}_{\psi}(\mathbf{z}_t,t,\mathbf{c})$ under conditioning $\mathbf{c}$ by matching the constant velocity along linear interpolants:
\begin{equation}
\mathbf{z}_0\sim\mathcal{N}(\mathbf{0},\mathbf{I}),\quad t\sim\mathcal{U}(0,1),\quad \mathbf{z}_t=(1-t)\mathbf{z}_0+t\mathbf{z}_1,
\end{equation}
\begin{equation}
\mathcal{L}_{\mathrm{cfm}}
=\mathbb{E}\Big[\big\|\mathbf{v}_{\psi}(\mathbf{z}_t,t,\mathbf{c})-(\mathbf{z}_1-\mathbf{z}_0)\big\|_2^2\Big].
\end{equation}
This latent generative pipeline enables efficient synthesis of geometry- and texture-consistent 3D assets, 
and the resulting structured latents capture rich joint statistics of shape and appearance, providing a strong transferable prior for fine-grained 3D part segmentation.

\subsection{Task Reformulation and I/O Representation}
\label{sec:task_define}

% We reformulate 3D part segmentation as a color prediction problem in a structured 3D representation.
% %
% This choice matches our base 3D generative model~\cite{trellis2}, which jointly parameterizes geometry together with appearance attributes such as color, material properties, and roughness in a unified representation.
% %
% To maximize reuse of the pretrained generative prior, 
% we avoid introducing an additional segmentation-specific attribute channel, which would increase modeling and optimization complexity, and instead express segmentation targets directly in color space, the most visually intuitive attribute.
% We consider three task settings with consistent input and output formats.

%
\textbf{Interactive part-segmentation} is formulated as binary part extraction: given user-provided 3D points indicating a target part, we supervise the model to color the selected part in white and the remaining regions in black.
\textbf{Full segmentation} targets multi-part decomposition: we assign each part a distinct color from a randomly sampled color palette and supervise voxel colors accordingly.
%
% Importantly, correctness is defined up to a permutation of colors within each object; any one-to-one assignment between predicted colors and parts is considered valid.
%
To reduce sensitivity to particular color choices, we use $K{=}10$ independently sampled palettes per shape, providing multiple colorizations for the same underlying partition.
\textbf{Full segmentation with 2D guidance} additionally conditions the model on a rendered 2D segmentation map: we first colorize the 3D parts and render the corresponding 2D segmentation map, and we then train the model to generate 3D voxel colors that are consistent with the color assignments in the 2D guidance.
Overall, this formulation preserves a unified model interface across settings, enabling a consistent architecture and training pipeline.

\subsection{Unified Multi-Task 3D Part Segmentation}
\label{sec:pipeline}

\subsubsection{Overall framework.}

To fully leverage pretrained 3D generative models, we cast 3D part segmentation as a conditional part-wise colorization task in 3D latent space.
Given an input asset $X$, a pretrained 3D VAE encoder $E(\cdot)$ produces a encoded latent $z = E(X)$, 
which helps specify the active voxel support and anchors generation to the underlying shape.
For each task, we construct a part-wise colorized target and encode it into the same latent space to obtain $y$, following the task-specific scheme in Sec.~\ref{sec:task_define}.
We then sample $\epsilon \sim \mathcal{N}(0,I)$ and $t\sim\mathcal{U}(0,1)$ to form a noisy interpolation
\begin{equation}
y_t \;=\; (1-t)\,y \;+\; t\,\epsilon .
\end{equation}
A pretrained DiT-based backbone is fine-tuned to predict the noise residual conditioned on the noisy input $y_t$, the geometry latent $z$, the task condition $C$, and a learned task embedding $e_{\tau}$:
\begin{equation}
\hat{v}_\theta \;=\; f_\theta\!\left(y_t,\, z,\, C,\, e_{\tau},\, t\right).
\end{equation}
Training follows the conditional flow-matching objective
\begin{equation}
\mathcal{L}(\theta)
\;=\;
\mathbb{E}_{X,\tau,t,\epsilon}\Big[
w(t)\,\big\| \hat{v}_\theta \;-\; (\epsilon - y) \big\|_2^2
\Big],
\end{equation}
where $w(t)$ is an optional timestep weighting.

\subsubsection{Condition Injection.}

We adopt task-specific conditioning designs while maintaining a unified interface across settings.
For interactive segmentation, user clicks in the UI provide an efficient and intuitive form of guidance.
In our framework, each click is encoded as a sparse point token comprising its 3D coordinates and an associated feature vector.
Since the 3D coordinates are already effectively encoded by RoPE within the attention layers, 
we omit the additional learnable input-level positional embedding used in prior designs~\cite{p3sam}.
Instead, all points share the same learnable feature vector $Q$ , which serves as the point token during both training and inference.
Given point coordinates $\mathbf{u}=\{u_i\}_{i=1}^{m}$ with $u_i\in\mathbb{R}^3$, we form point-condition tokens
\begin{equation}
Q \;=\; \big[\; \mathbf{q}(u_1),\ldots,\mathbf{q}(u_m) \;\big], 
\qquad
\mathbf{q}(u_i) \;=\; \big[\, u_i \,;\, \mathbf{e}_p \,\big],
\end{equation}
where $\mathbf{e}_p$ is a shared learnable feature appended to every point token.
Conditioned on $Q$, the denoising model is instantiated as
\begin{equation}
\hat{v}_{\theta} \;=\; f_{\theta}\!\left(y_t,\, z,\, Q,\, e_{\tau},\, t\right).
\end{equation}
When the number of points is fewer than $10$, we pad the point tokens to a length of $10$
using zero coordinates and zero features.
To preserve a single unified model, we keep this interface for full segmentation and 2D-guided full segmentation by providing $10$ padded tokens with all-zero coordinates and features.

For full segmentation with 2D guidance, we additionally provide a user-specified 2D segmentation colorization as guidance. In this setting, the guidance specifies the desired part decomposition in image space, which is then transferred to 3D through our generative framework. This provides an explicit way to obtain finer or coarser 3D segmentations when such decomposition is indicated by the input 2D map, while interactive segmentation further supports practical refinement by extracting and merging local regions through additional user clicks.
The guidance image is encoded into a sequence of conditioning tokens injected via cross-attention:
\begin{equation}
p \;=\; g_{\phi}(I_{\text{guide}}),
\end{equation}
where $g_{\phi}(\cdot)$ denotes an image encoder.
In this setting, denoising is conditioned on both the padded point-token interface $Q_{0}$ and the image guidance tokens $p$:
\begin{equation}
\hat{v}_{\theta} \;=\; f_{\theta}\!\left(y_t,\, z,\, (Q_{0},\, p),\, e_{\tau},\, t\right).
\end{equation}

\begin{table*}[t]
\small
\centering
\caption{Comparison of interactive part segmentation performance. We report IoU at different numbers of clicks, compared with Point-SAM~\cite{pointsam} and P3-SAM~\cite{p3sam} on PartObjaverse-Tiny~\cite{partobjaverse_tiny} and PartNeXT~\cite{partnext}.}
\label{tab:click_segmentation}
\begin{tabular}{l|ccccc ccccc}
\toprule
\multirow{2}{*}{Method}
& \multicolumn{5}{c}{PartObjaverse-Tiny}
& \multicolumn{5}{c}{PartNeXT} \\
\cmidrule(lr){2-6}\cmidrule(lr){7-11}
& IoU@1 & IoU@3 & IoU@5 & IoU@7 & IoU@10
& IoU@1 & IoU@3 & IoU@5 & IoU@7 & IoU@10 \\
\midrule
Point-SAM~\cite{pointsam} & 24.87 & 48.99 & 59.67 & 64.33 & 67.99 & 23.90 & 47.50 & 56.71 & 61.23 & 65.04 \\
P3-SAM~\cite{p3sam}       & 33.04 & 50.57 & 53.78 & 54.74 & 55.51 & 35.61 & 51.26 & 52.03 & 52.61 & 53.81 \\
\textbf{SegviGen}         & \textbf{42.49} & \textbf{61.14} & \textbf{67.53} & \textbf{71.50} & \textbf{75.02}
                          & \textbf{54.86} & \textbf{71.15} & \textbf{78.11} & \textbf{79.96} & \textbf{82.73} \\
\bottomrule
\end{tabular}
\end{table*}

\begin{figure*}
  \centering
  \includegraphics[width=\linewidth]{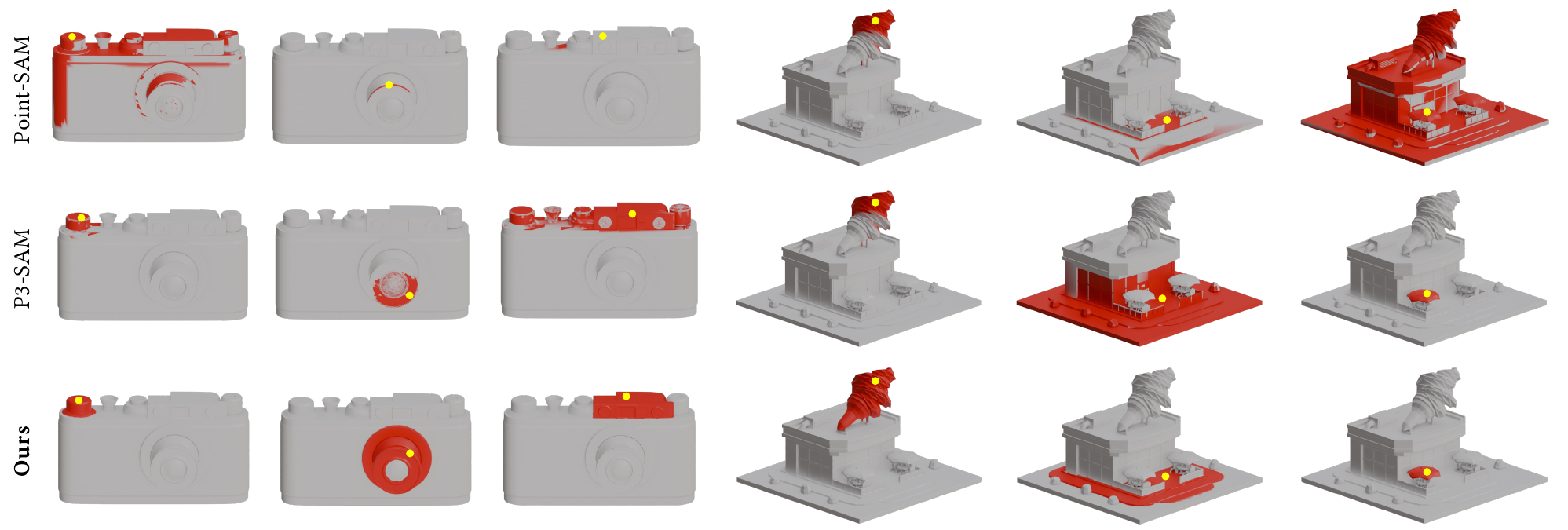}
  \caption{
  Interactive part-segmentation results. 
  We compare \textit{SegviGen} with existing representative baselines, including Point-SAM~\cite{pointsam} and P3-SAM~\cite{p3sam}. 
  In the figure, yellow points denote user clicks, while the predicted target part is highlighted in red.
  Leveraging priors from pretrained 3D generative models, \textit{SegviGen} achieves more accurate results with sharper boundaries than prior methods, while requiring substantially less training data.
  }
  \label{fig:comparison_point}
\end{figure*}

\subsubsection{Task Embedding.}

To improve multi-task generalization within a single model, task identity is encoded as a continuous embedding and injected alongside the timestep signal.
Let $\tau\in\{1,\dots,T\}$ denote the task index.
A sinusoidal encoding is first computed from $\tau$,
\begin{equation}
s_{\tau} \;=\; \mathrm{PE}(\tau)\in\mathbb{R}^{d_f},
\end{equation}
where $\mathrm{PE}(\cdot)$ follows the standard sinusoidal scheme.
A lightweight MLP then maps $s_{\tau}$ to the task embedding
\begin{equation}
e_{\tau} \;=\; \mathrm{MLP}_{\psi}(s_{\tau}) \in \mathbb{R}^{d}.
\end{equation}
In parallel, the timestep $t$ is embedded as $e_t\in\mathbb{R}^{d}$.
The final modulation vector used by DiT backbone is obtained by additive fusion,
\begin{equation}
m \;=\; e_t \;+\; e_{\tau},
\qquad
\hat{v}_{\theta} \;=\; f_{\theta}\!\left(y_t,\, z,\, C,\, m\right),
\end{equation}
where $m$ conditions the adaptive layers to jointly encode diffusion progress and task semantics.
During training, samples from different tasks are interleaved and supervised with their corresponding $\tau$, 
encouraging the shared backbone to learn task-discriminative behaviors while preserving a unified parameterization.

\section{EXPERIMENTS}
\subsection{Setting}
\paragraph{Implementation Details.}

\begin{figure*}[t]
  \centering
  \includegraphics[width=\linewidth]{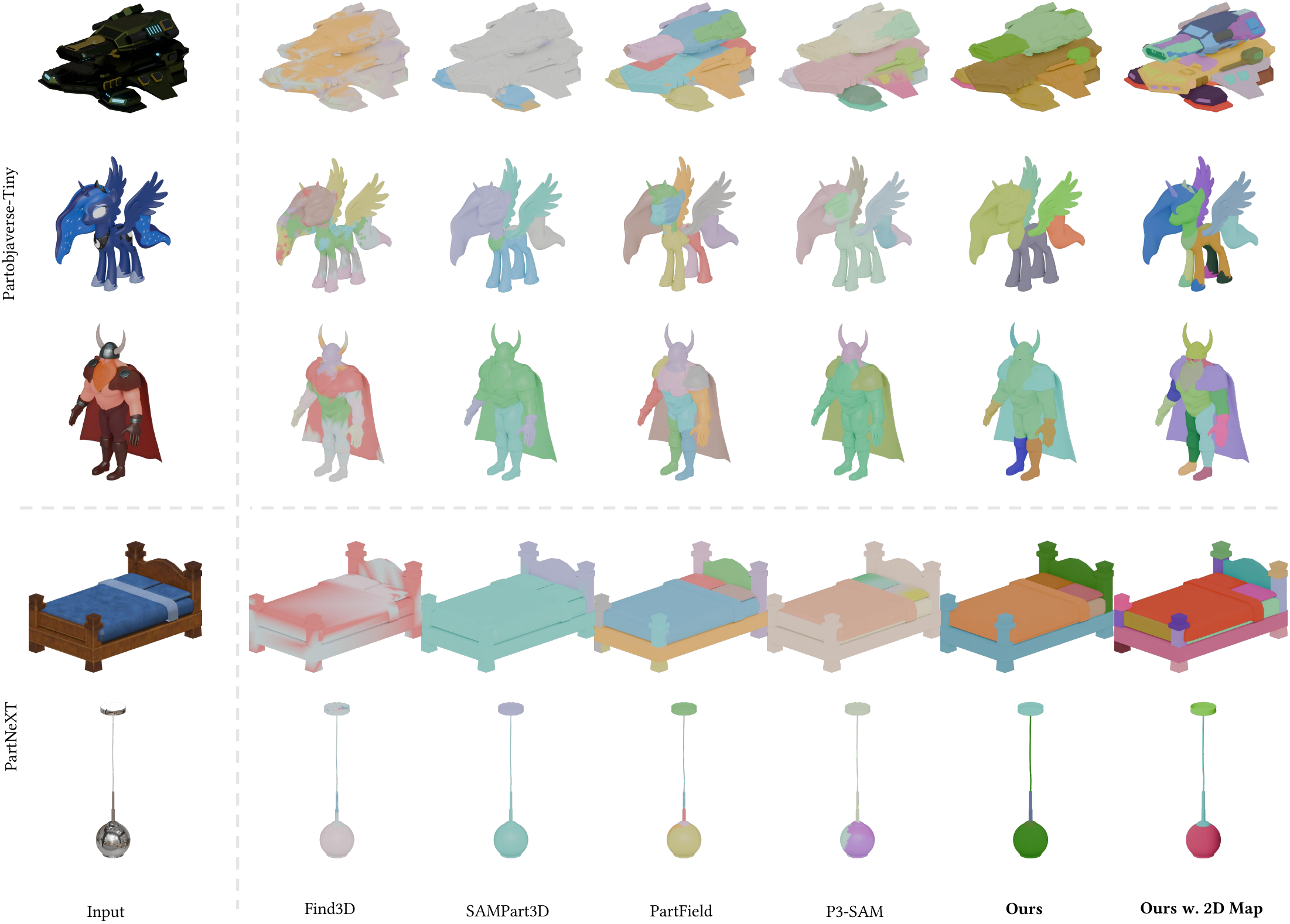}
  \caption{
    Full segmentation results. 
    We compare \textit{SegviGen} against a broad set of prior methods, where different colors indicate different segmented parts. 
    From the results, \textit{SegviGen} achieves high-accuracy full segmentation with sharp part boundaries using only 3D input.
  }
  \label{fig:comparison_full}
\end{figure*}

We adopt Trellis.2~\cite{trellis2} as our base model, which is a 3D generative framework with a native and compact structured latent representation. 
For all experiments, the Tex-SLAT flow model is trainable, while the remaining SC-VAE is kept frozen.
We adopt the AdamW optimizer~\cite{adamw} with a learning rate of $1\times10^{-4}$. All experiments are conducted on 8 NVIDIA A800 GPUs, and the model is trained for  8 hours.
Unless otherwise specified, the segmentation results shown in this paper are produced with 12-step inference.

\paragraph{Datasets.}
For training, we use the PartVerse dataset~\cite{copartpartverse}, which contains 12k objects with a total of approximately 91k annotated parts.
For evaluation, we use PartObjaverse-Tiny~\cite{sampart3d}, which contains 200 textured mesh objects, and a 300-object textured-mesh subset of PartNeXT~\cite{partnext}.

\paragraph{Baselines.}
We compared our model's performance on full segmentation between P3-SAM~\cite{p3sam}, Find3D~\cite{find3d}, SAMPart3D~\cite{sampart3d}, Partfield~\cite{partfield}.
P3-SAM is a native 3D point-promptable part segmenter with multiple mask heads and an IoU predictor.
% It can be run automatically by sampling prompt points and merging redundant masks with NMS.
Find3D targets open-world, language-queryable parts by auto-labeling rendered multi-view images with SAM and a VLM.
% , projecting them back to 3D, and training a transformer to produce per-point features aligned to a CLIP-like embedding space for cosine-similarity querying.
SAMPart3D and PartField both learn part-aware 3D features from multi-view SAM masks and obtain parts via feature clustering.

For interactive part segmentation, we compared our model against P3-SAM~\cite{p3sam} and Point-SAM~\cite{pointsam}, where Point-SAM adapts the SAM prompt-and-mask paradigm to point clouds and is trained with SAM-generated pseudo masks.
\paragraph{Metrics.}
To evaluate the interactive segmentation, we sample 10 positive points for each part, then measure the average IOU between the predicted masks for all clicks of all parts and their corresponding ground truth masks. \textbf{IoU@N} stands for IoU score in \textbf{N} foreground clicks.
The evaluation metric for full segmentation is the same method in previous work~\cite{p3sam,partfield}, using IoU to measure the accuracy of overall mask predictions. 

\paragraph{Voxel-to-Mesh Color Transfer.} \textit{SegviGen} predicts part-indicative colors on active O-Voxels. Since the mesh decoded by Trellis.2 may differ from the input in tessellation and local topology, we transfer the predicted voxel colors back to the original input mesh. Specifically, each mesh vertex is assigned the color of its nearest active voxel, and each face label is determined by majority voting over its vertices. This preserves the original mesh structure and is more suitable for mesh-level segmentation than directly using the decoded mesh. We further apply lightweight mesh-level smoothing to remove isolated spikes introduced by projection near part boundaries.
% 1. Setting

% 1.1 Implementation details
% 1.2 Baselines
% 1.3 Metrics
\subsection{Main Results}
\subsubsection{Interactive Part-Segmentation}

We evaluate interactive part segmentation on two benchmarks: PartObjaverse-Tiny~\cite{partobjaverse_tiny} and PartNeXT~\cite{partnext}.
We benchmark against two state-of-the-art native 3D methods: Point-SAM~\cite{pointsam}, which is specialized for point cloud segmentation, and P3-SAM~\cite{p3sam}. The quantitative results are summarized in Table~\ref{tab:click_segmentation}.

As shown in Table~\ref{tab:click_segmentation}, \textit{SegviGen} consistently outperforms all baselines by a significant margin across all interaction rounds. Notably, our method demonstrates exceptional efficiency in the \textit{few-shot} interaction setting. In the most challenging $1$-click scenario (IoU@1), \textit{SegviGen} achieves \textbf{42.49\%} on PartObjaverse-Tiny and \textbf{54.86\%} on PartNext, surpassing the Point-SAM by approximately \textbf{17.6\%} and \textbf{31.0\%}, respectively. This indicates that our generative framework possesses a much stronger initial understanding of 3D part structures compared to discriminative approaches, allowing it to infer complete part geometries from minimal user guidance.

Furthermore, as the number of user clicks increases from 1 to 10, \textit{SegviGen} exhibits a steady and robust performance gain. On the PartNext dataset, our method reaches an IoU of \textbf{82.73\%} at 10 clicks, significantly higher than Point-SAM (65.04\%) and P3-SAM (53.81\%). This demonstrates that our model effectively incorporates user feedback to refine boundaries and resolve ambiguities.

\subsubsection{Full Segmentation}

We evaluate the full segmentation capability of \textit{SegviGen} in two distinct settings:
(1) Using purely native 3D representation.
(2) Incorporating with 2D guidance. 
Quantitative comparisons with state-of-the-art methods, including Find3D~\cite{find3d}, SAMPart3D~\cite{sampart3d}, PartField~\cite{partfield}, and P3-SAM~\cite{p3sam}, are presented in Table~\ref{tab:auto_segmentation}.
Qualitative results are shown in \ref{fig:comparison_full}.

\paragraph{Without 2D Guidance}
In this setting, \textit{SegviGen} performs segmentation solely based on the structural and appearance priors learned during pretraining, without access to any external 2D segmentation maps. The model is prompted to generate part-indicative colors directly from the latent 3D representation.
As shown in Table~\ref{tab:auto_segmentation}, our method demonstrates superior generalization, particularly on PartNext. \textit{SegviGen} achieves an IoU of \textbf{55.40\%}, significantly outperforming PartField (41.50\%) and SAMPart3D (29.62\%)
While SAMPart3D performs well on the smaller PartObjaverse-Tiny dataset (59.05\%), its performance collapses on PartNext. In contrast, \textit{SegviGen} maintains robust performance (50.64\% on PartObjaverse-Tiny).

\paragraph{With 2D Guidance}
To further unleash the potential of \textit{SegviGen}, we introduce a 2D-guided mode where the model is conditioned on a single-view 2D segmentation map (rendered via nvdiffrast or derived from a 2D segmenter). This setting combines the rich semantic cues of 2D foundation models with the geometric consistency of our 3D generative framework.
Incorporating this lightweight 2D prior yields substantial performance gains. As shown in Table~\ref{tab:auto_segmentation}, \textit{SegviGen} (w. 2D Map) achieves new state-of-the-art results on both datasets, reaching \textbf{62.98\%} on PartObjaverse-Tiny and \textbf{71.53\%} on PartNext.

\begin{figure}[t]
  \centering
  \includegraphics[width=\linewidth]{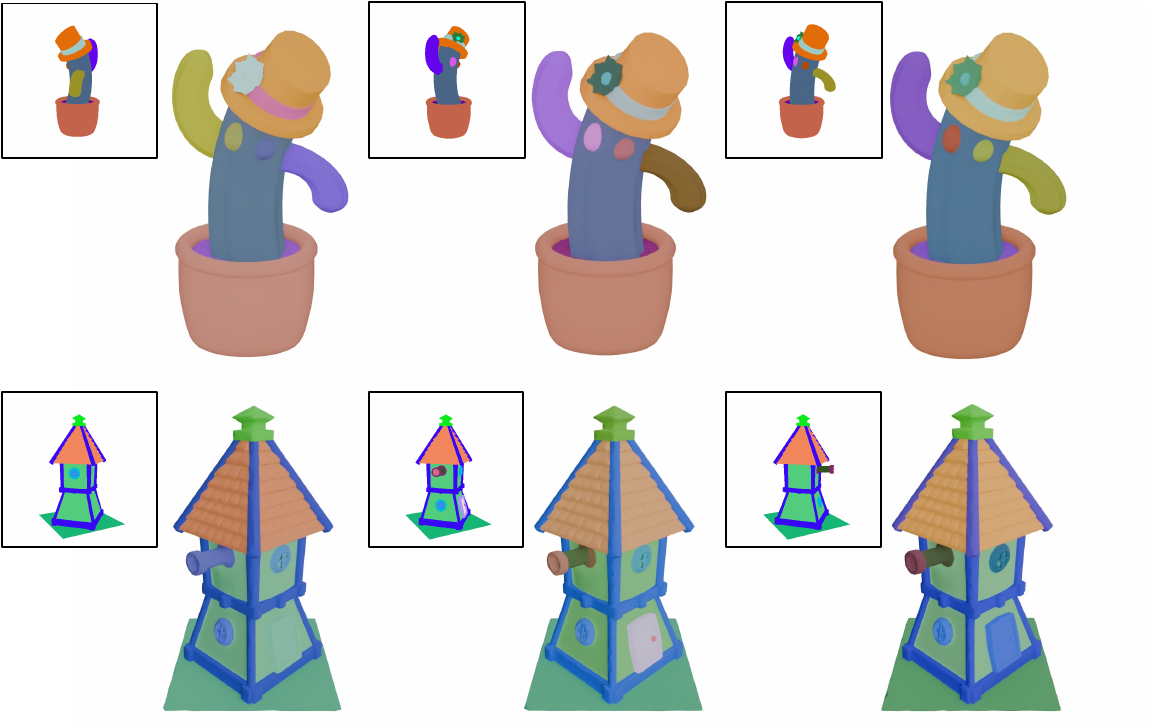}
  \caption{
    Results with different view 2D guidance.
    Visible regions are consistently segmented, while occluded regions may receive different color assignments under different views.
  }
  \label{fig:different_view}
\end{figure}

\subsubsection{Effect of 2D Segmentation Guidance}
We further evaluate \textit{SegviGen} with 2D segmentation maps rendered from different viewpoints. As shown in \ref{fig:different_view}, visible regions in the guidance map are reliably transferred to 3D. For regions invisible from the guided view, different viewpoints may lead to different color assignments. However, these differences mainly reflect label-assignment ambiguity rather than incorrect decomposition, as the resulting parts remain consistent and plausible. This suggests that \textit{SegviGen} can effectively absorb different 2D segmentation results as guidance, while the exact color assignment of occluded regions may vary across views.

\begin{figure}[t]
  \centering
  \includegraphics[width=0.85\linewidth]{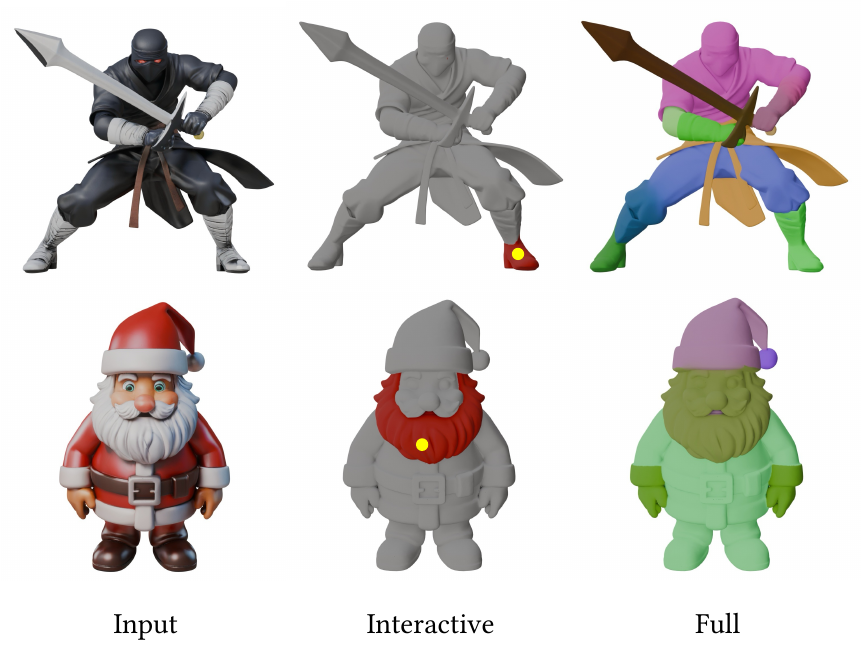}
  \caption{
    Generalization results on AI-generated meshes.
    The results show that \textit{SegviGen} can produce plausible interactive and full part segmentations on AI-generated 3D assets without additional training.
  }
  \label{fig:generalization_results}
\end{figure}

\subsubsection{Generalization to AI-Generated Meshes}
To evaluate generalization beyond artist-created meshes, we further test \textit{SegviGen} on meshes generated by Hunyuan3D 2.1. These meshes differ from the training and benchmark assets in geometry quality, topology, and part composition, and do not have ground-truth part annotations. We therefore provide qualitative results in \ref{fig:generalization_results}. The results show that \textit{SegviGen} can produce plausible part decompositions on AI-generated 3D assets, demonstrating its potential applicability to automatically generated 3D content.

\begin{table}[t]
\small
  \centering
  \caption{Quantitative results (IoU) for full segmentation. “w. 2D Map” denotes the setting with 2D segmentation-map guidance.}
  \label{tab:auto_segmentation}
  \setlength{\tabcolsep}{8pt}
  \begin{tabular}{l|cc}
    \toprule
    Method & PartObjaverse-Tiny & PartNext~ \\
    \midrule
    Find3D~\cite{find3d}            & 15.62 & 19.04 \\
    SAMPart3D~\cite{sampart3d}        & 59.05 & 29.62 \\
    PartField~\cite{partfield}       & 51.72 & 41.50 \\
    P3-SAM~\cite{p3sam}            & 45.36 & 31.94 \\
    \hline
    \textbf{SegviGen} & \textbf{50.64} & \textbf{55.40} \\
    \textbf{SegviGen (w. 2D Map)} & \textbf{62.98} & \textbf{71.53} \\
    \bottomrule
  \end{tabular}
\end{table}

\begin{figure*}[t]
  \centering
  \includegraphics[width=0.85\linewidth]{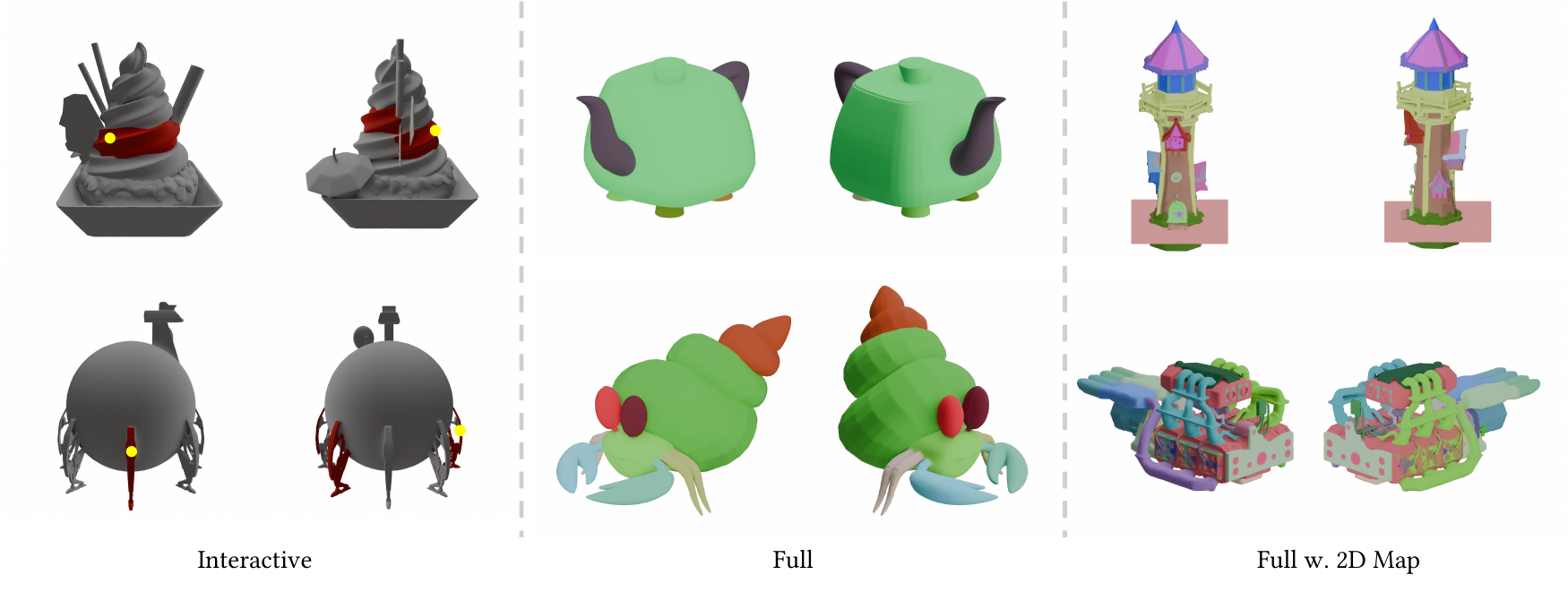}
  \caption{
    Failure cases.
    Due to semantic ambiguity, \textit{SegviGen} may produce slightly more or fewer parts than expected in interactive and full segmentation. In the 2D-guided setting, extremely fine structures may still be difficult to segment clearly and consistently from the guidance map.
  }
  \label{fig:fail_cases}
\end{figure*}

% 表，图
\subsection{Ablation Studies and Analysis}
\subsubsection{Point Embedding Mechanism}
To investigate the optimal representation point prompt within our framework, we conducted an ablation study on the point embedding mechanism, comparing two distinct strategies:

\paragraph{Explicit Coordinate Encoding}In this setting, spatial coordinates are explicitly injected into the feature space. We utilize a frequency-based positional encoding scheme to map continuous 3D coordinates into high-dimensional embeddings, which will fuse with learnable semantic vectors. Consequently, the input features explicitly encapsulate both absolute spatial information and semantic category.

\paragraph{Label-based Semantic Embedding}In this setting, the feature vectors serve solely as semantic indicators without explicitly encoding geometric values. A shared learnable embedding vector is assigned to all foreground points. The spatial information is preserved implicitly via the coordinate indices of the SparseTensor, relying on the sparse backbone's intrinsic ability to process spatial locality. As shown in Table \ref{tab:ablation_embedding}, as the number of interactions increases, the Explicit Coordinate Encoding method outperforms the Label-based approach, particularly in the later stages.

\begin{table}[t]
\small
  \centering
  \caption{Ablation study on point embedding mechanisms. We compare the performance of Explicit Coordinate Encoding and Label-based Semantic Embedding under varying numbers of clicks on PartObjaverse~\cite{partobjaverse_tiny}.}
  \label{tab:ablation_embedding}
  \setlength{\tabcolsep}{6pt}
  \begin{tabular}{l|ccccc}
    \toprule
    Method & IoU@1 & IoU@3 & IoU@5 & IoU@7 & IoU@10 \\
    \midrule
    Explicit Coord & 41.75 & 60.19 & 67.43 & 71.61 & 75.40 \\
    Label-based    & \textbf{42.49} & \textbf{61.14} & 67.53 & 71.50 & 75.02 \\
    \bottomrule
  \end{tabular}
\end{table}

\begin{table}[t]
\small
    \centering
    \caption{Effect of sampling steps on segmentation performance. We choose 12 steps as a practical trade-off between accuracy and efficiency.}
    \label{tab:steps_analysis}
    \setlength{\tabcolsep}{5pt}
    \begin{tabular}{l|cccccc}
    \toprule
    Steps & IoU@1 & IoU@3 & IoU@5 & IoU@7 & IoU@10 & Time \\
    \midrule
    1 & 42.90 & 59.98 & 65.86 & 69.50 & 72.85 & 0.44s \\
    4 & \textbf{44.51} & 60.40 & 66.65 & 70.64 & 73.58 & 1.02s \\
    8 & 44.21 & 61.14 & 67.64 & 71.14 & 74.49 & 1.81s \\
    12 & 42.49 & 61.14 & 67.53 & 71.50 & \textbf{75.02} & 2.63s \\
    25 & 43.82 & \textbf{61.67} & \textbf{68.30} & \textbf{71.87} & 74.99 & 5.12s \\
    \bottomrule
    \end{tabular}
\end{table}

% 3.1 Point embedding 机制

\subsubsection{Number of denoising steps at inference.}

We analyze the impact of sampling steps on segmentation performance in Table \ref{tab:steps_analysis}. Due to the trajectory property of the flow model, we observe a great performance even with one step. Performance improves as steps increase, but gains begin to saturate over 8 steps. Although 25 steps offer marginal improvements, the inference latency nearly doubles compared to 12 steps. Thus we adopt 12 steps as the optimal balance between high-quality results and computational efficiency.

\subsection{Failure Cases and Limitations}
\textit{SegviGen} still has two main limitations. First, in the interactive and full segmentation settings, semantic ambiguity may cause the model to produce more or fewer parts than expected, since multiple valid part decompositions may exist for the same object. Second, although 2D guidance improves controllability, the model cannot always reproduce highly detailed part decompositions specified by the input 2D segmentation map. When the guidance becomes overly fine-grained, the resulting 3D segmentation may exhibit reduced boundary precision and smoothness. Addressing these issues with more explicit semantic control and fine-structure-aware segmentation remains an important direction for future work.

% 对比快

% 表

% Additional
% 更多分割结果（现有的3D模型，AI生成的3D模型，场景分割）% lin
% 更多和P3SAM的对比
% 交互式demo的截图 （优化）   % lin
% 3D 打印的图 % lin
% 更多应用，part-aware 3D generation, 3D 编辑

\section{CONCLUSION}

This paper introduces \textit{SegviGen}, a framework that repurposes pretrained 3D generative models for 3D part segmentation. 
In contrast to 2D-to-3D lifting methods that often suffer from cross-view inconsistency and blurred boundaries, 
and native 3D discriminative approaches that require large-scale part annotations and heavy training, 
\textit{SegviGen} transfers generative priors to deliver accurate and globally coherent segmentations with limited supervision. 
It reformulates segmentation as part-wise colorization, jointly reconstructing geometry and predicting part-indicative colors, 
and supports multiple task settings via flexible conditioning. 
Experiments on interactive and full segmentation benchmarks show consistent improvements over prior methods, underscoring the effectiveness and data efficiency of 3D generative priors for 3D part segmentation.

\section*{Acknowledgment}
This work was supported by National Natural Science Foundation of China (62132001), Beijing Natural Science Foundation (L252218), and the Fundamental Research Funds for the Central Universities.

\bibliographystyle{ACM-Reference-Format}
\bibliography{sample-bibliography}

\newpage

\begin{figure*}[t]
  \centering

  % Row 1
  \begin{minipage}[t]{0.49\linewidth}
    \centering
    \includegraphics[width=\linewidth]{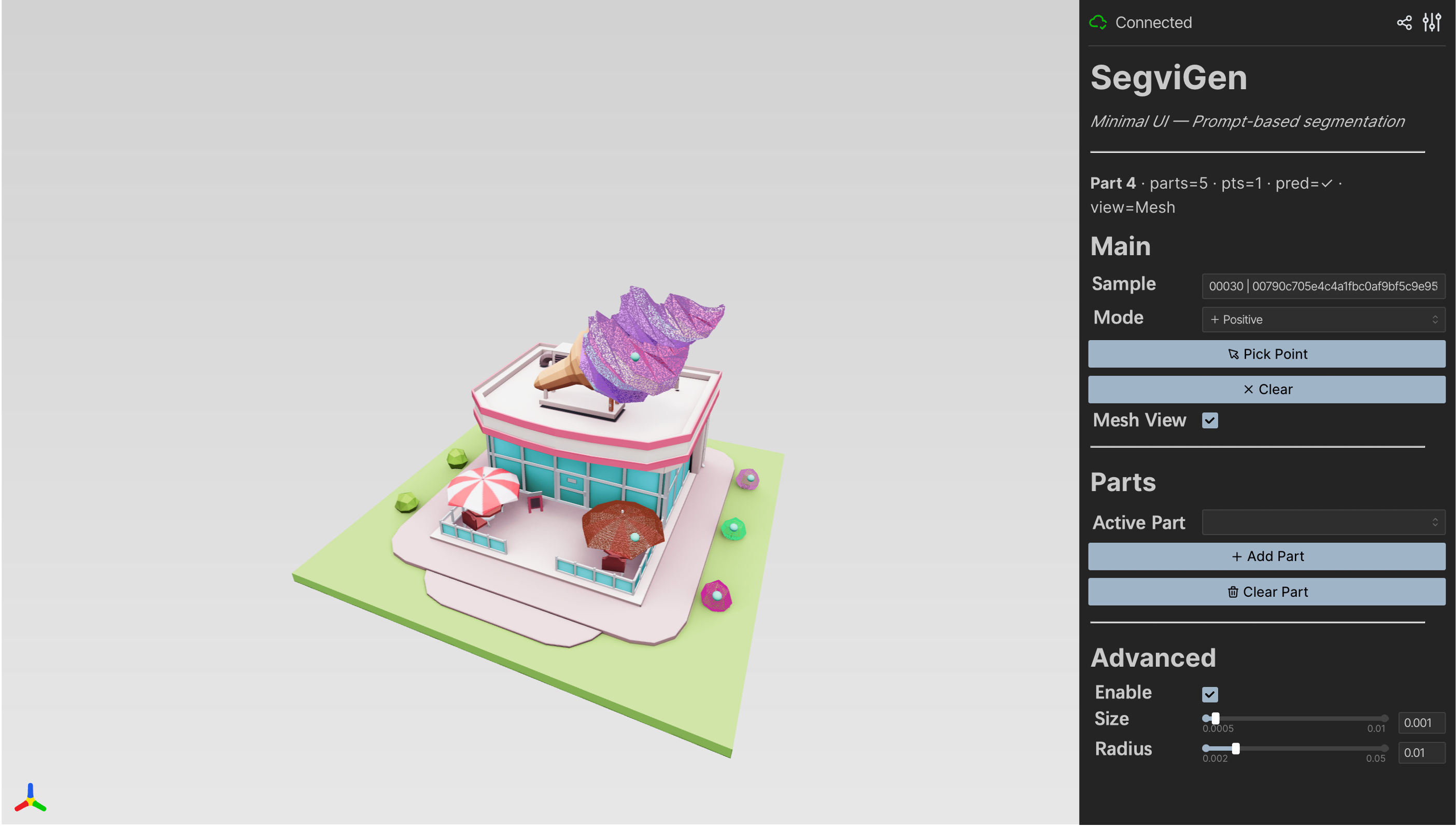}\\[-0.2em]
    % {\small (a)}
  \end{minipage}
  \hspace{0.01\linewidth}
  \begin{minipage}[t]{0.49\linewidth}
    \centering
    \includegraphics[width=\linewidth]{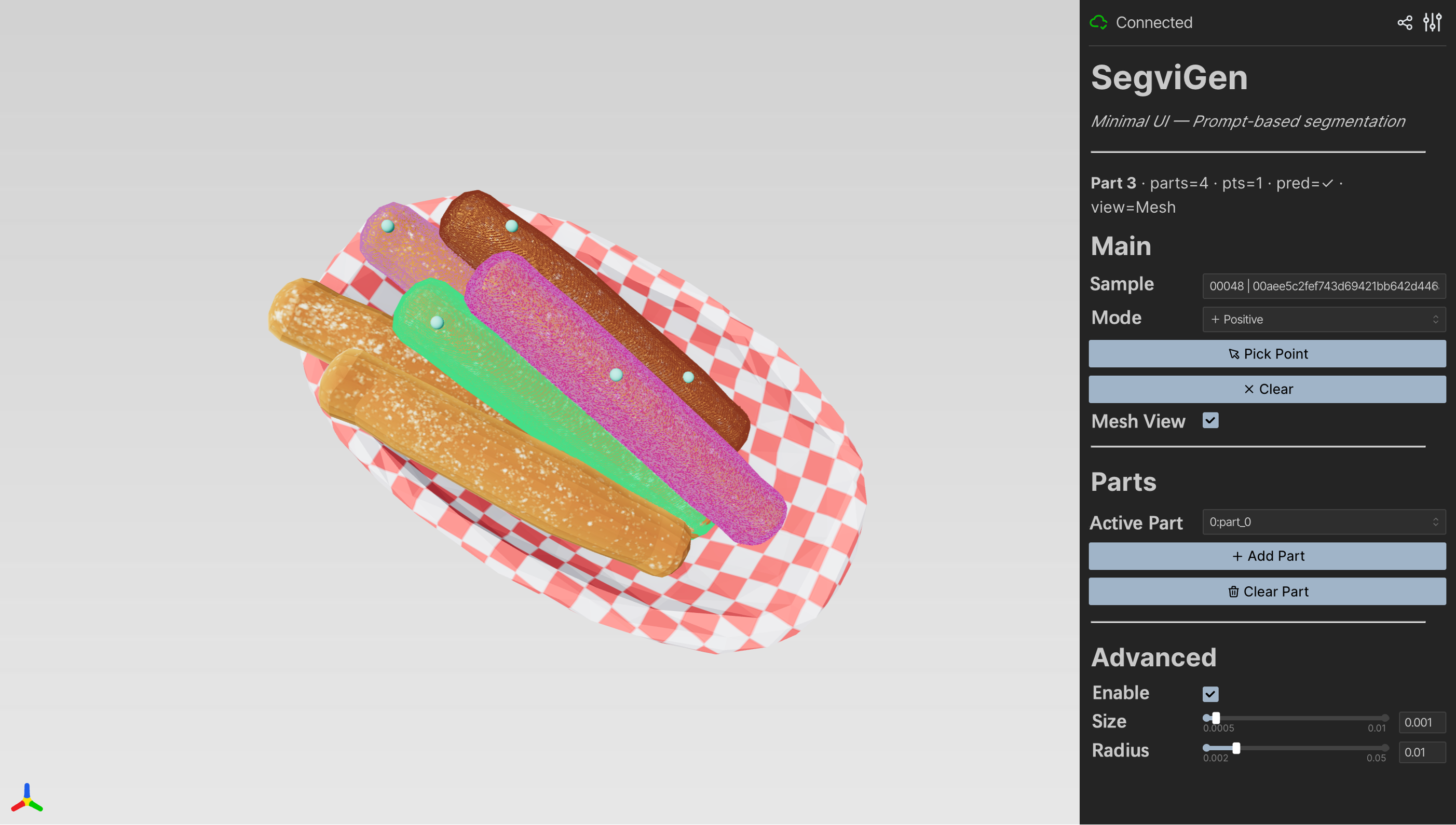}\\[-0.2em]
    % {\small (b)}
  \end{minipage}

  \vspace{0.01\linewidth}

  % Row 2
  \begin{minipage}[t]{0.49\linewidth}
    \centering
    \includegraphics[width=\linewidth]{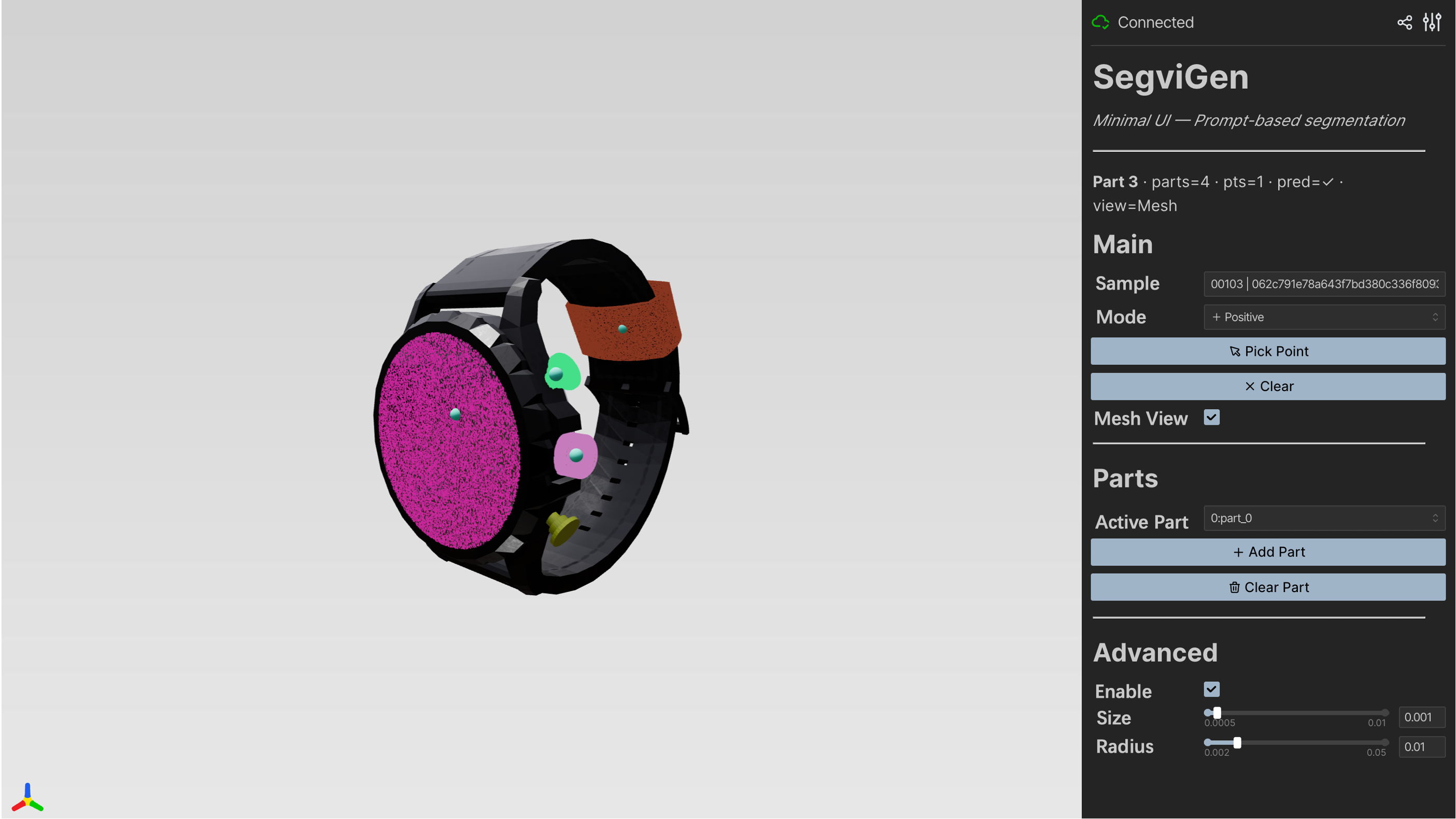}\\[-0.2em]
    % {\small (c)}
  \end{minipage}
  \hspace{0.01\linewidth}
  \begin{minipage}[t]{0.49\linewidth}
    \centering
    \includegraphics[width=\linewidth]{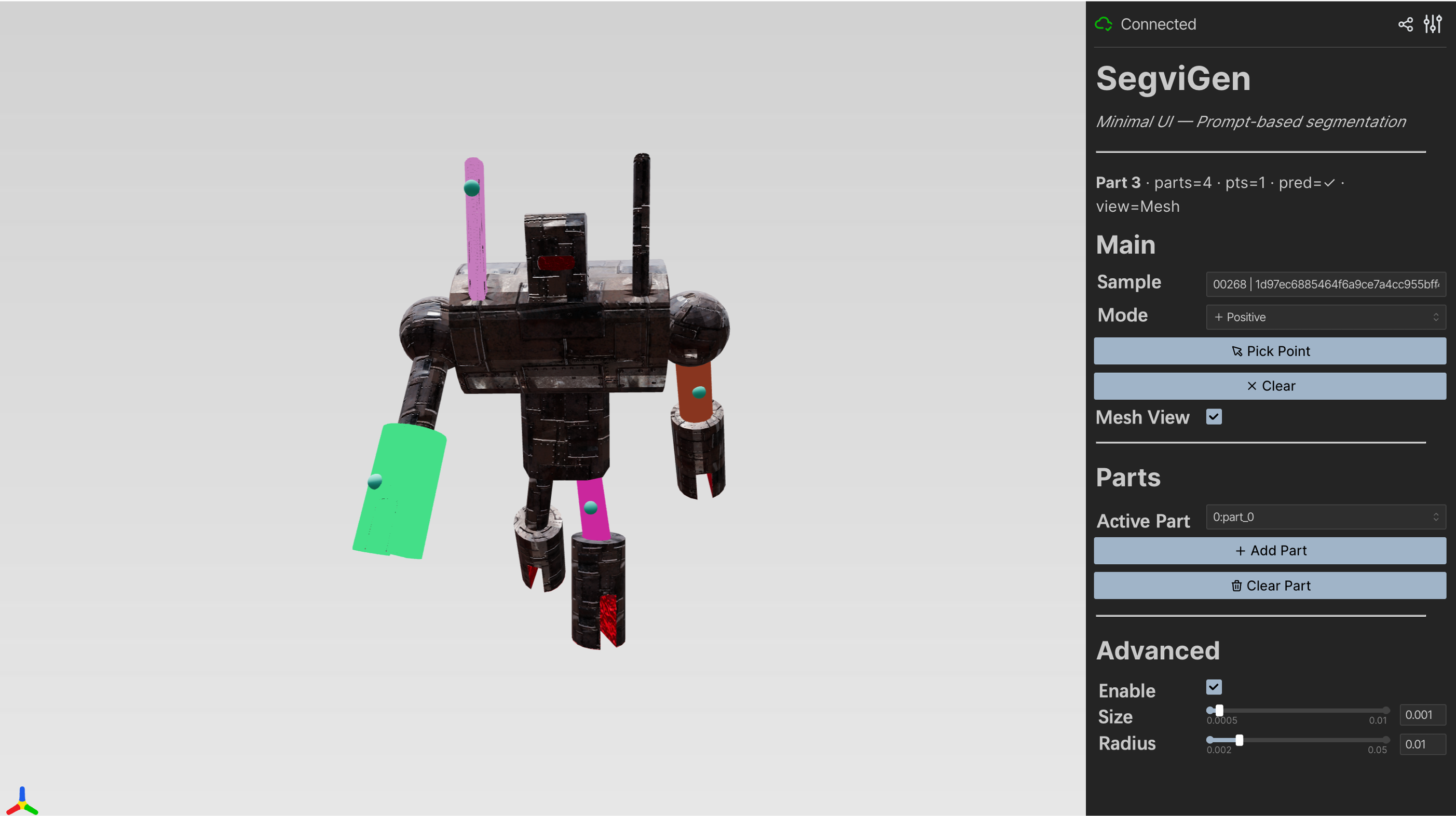}\\[-0.2em]
    % {\small (d)}
  \end{minipage}

  \caption{Interactive demo. Users specify clicks on the 3D asset to perform interactive part segmentation, and can adjust visualization settings.}
  \label{fig:web_screenshot}
\end{figure*}

\begin{figure*}
  \centering
  \includegraphics[width=\linewidth]{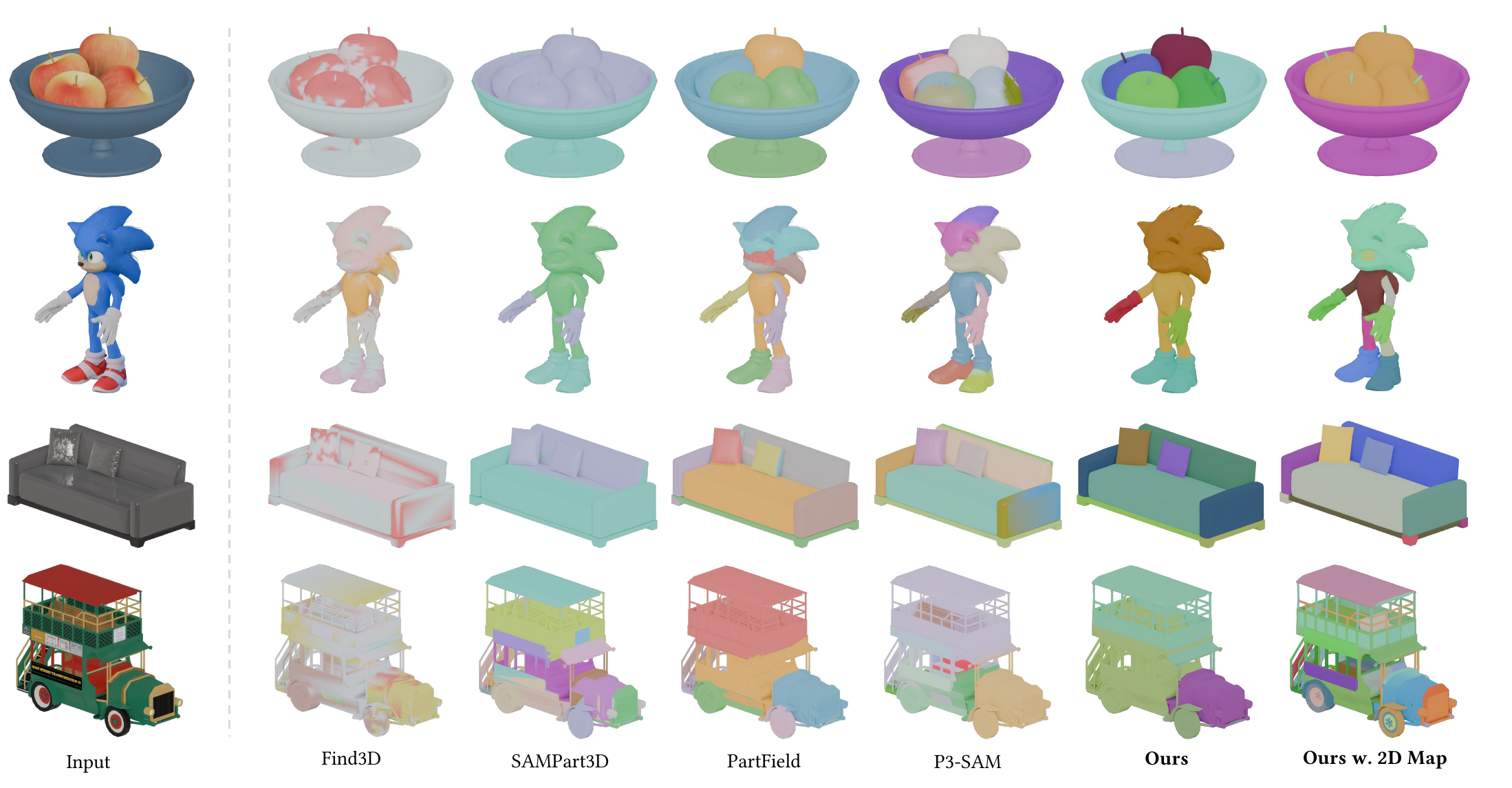}
  \caption{More qualitative comparisons for full segmentation.}
  \label{fig:comparison_more_result}
\end{figure*}

\begin{figure*}
  \centering
  \includegraphics[width=0.92\linewidth]{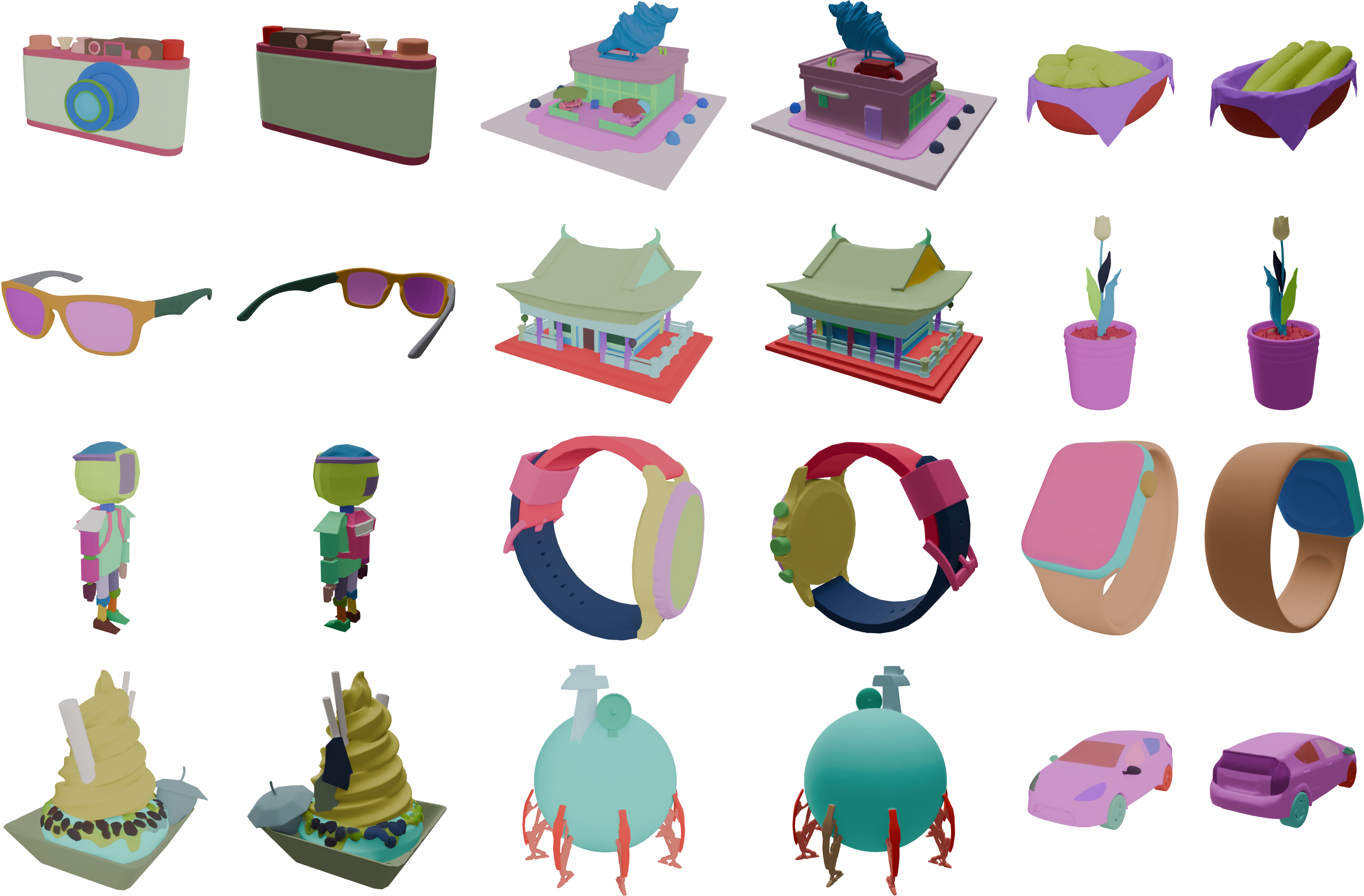}
  \caption{More qualitative segmentation results of our \textit{SegviGen}.}
  \label{fig:more_results_ours}
\end{figure*}

\begin{figure*}
  \centering
  \includegraphics[width=0.92\linewidth]{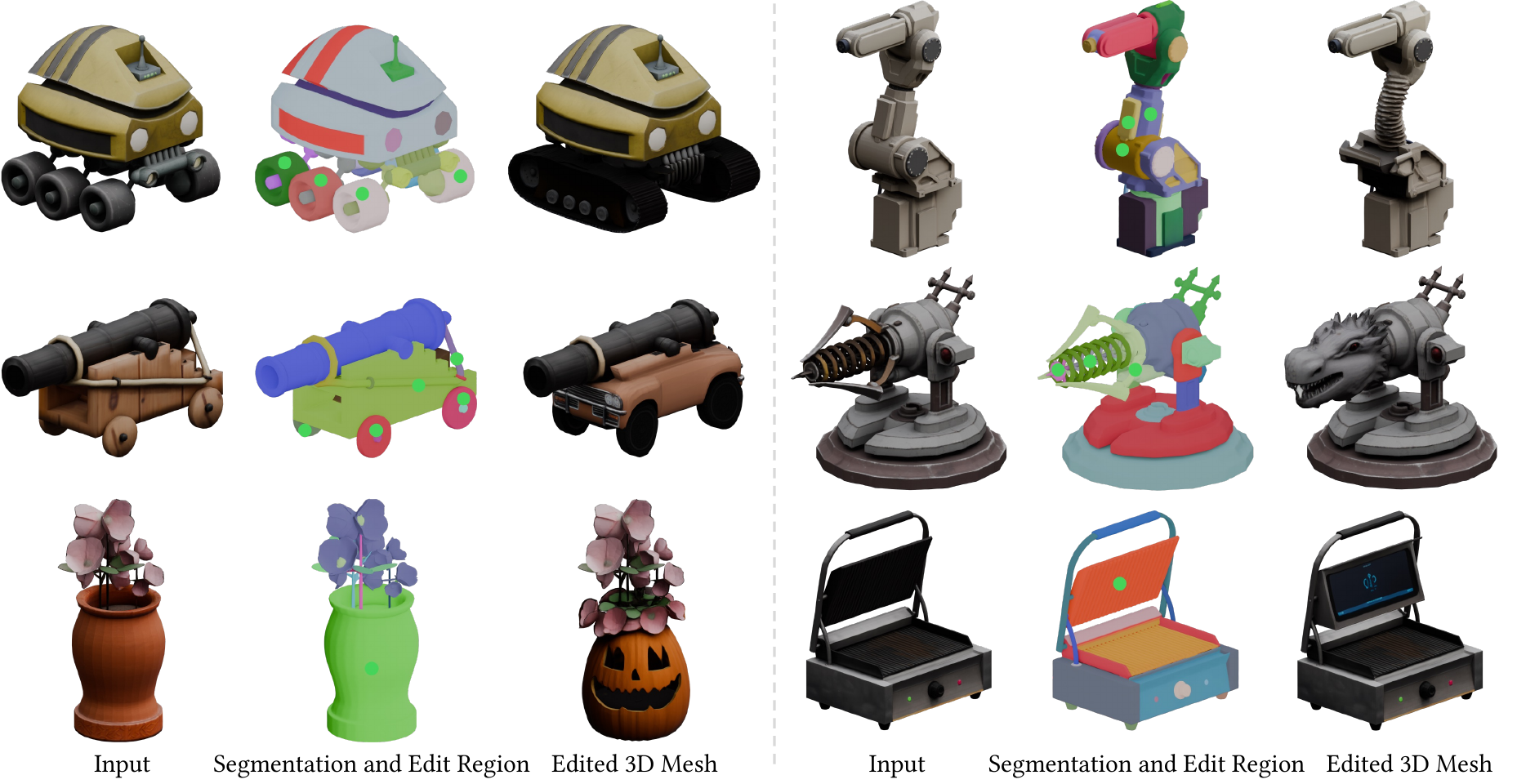}
  \caption{
    Interactive 3D editing results with \textit{SegviGen} and \textit{VoxHammer}~\cite{voxhammer}.
    \textit{SegviGen} provides precise part segmentation to facilitate downstream editing models. 
    The target region to be edited is indicated by green points.
    It demonstrates the practical utility of \textit{SegviGen} in downstream 3D editing pipelines.
    }
  \label{fig:more_apps_edit}
\end{figure*}

  \end{document}